# Attention-Driven Multi-Agent Reinforcement Learning: Enhancing Decisions with Expertise-Informed Tasks

**Andre R. Kuroswiski[1,2], Annie S. Wu[1], Angelo Passaro[3]**

[1]University of Central Florida, Orlando, FL 32816-2362, USA
[2]Aeronautics Institute of Technology, São José dos Campos, SP 12228-900, Brazil
[3]Institute for Advanced Studies, São José dos Campos, SP 12228-001, Brazil
`kuroswiski@ita.br, aswu@cs.ucf.edu, angelo@ieav.cta.br`

## Abstract

In this paper, we introduce an alternative approach to enhancing Multi-Agent Reinforcement Learning (MARL) through the integration of domain knowledge and attention-based policy mechanisms. Our methodology focuses on the incorporation of domain-specific expertise into the learning process, which simplifies the development of collaborative behaviors. This approach aims to reduce the complexity and learning overhead typically associated with MARL by enabling agents to concentrate on essential aspects of complex tasks, thus optimizing the learning curve. The utilization of attention mechanisms plays a key role in our model. It allows for the effective processing of dynamic context data and nuanced agent interactions, leading to more refined decision-making. Applied in standard MARL scenarios, such as the Stanford Intelligent Systems Laboratory (SISL) Pursuit and Multi-Particle Environments (MPE) Simple Spread, our method has been shown to improve both learning efficiency and the effectiveness of collaborative behaviors. The results indicate that our attention-based approach can be a viable approach for improving the efficiency of MARL training process, integrating domain-specific knowledge at the action level.

## Introduction

This work proposes an alternative methodology designed to simplify the Multi-Agent Reinforcement Learning (MARL) process by incorporating previous knowledge about the problem at the agent action level. Our approach aims to reduce the learning overhead through predefined task-based actions enriched with domain knowledge and leverages the ability of attention mechanisms to handle variable and complex contexts to enhance collaborative behaviors.

MARL is a subfield of AI that aims to develop algorithms for training multiple agents to interact with each other and their environment, learning optimal decision-making policies through trial and error. MARL is particularly challenging as agents' actions can affect each other's learning process and the overall state of the environment, requiring the development of sophisticated techniques to promote coordination and adaptation in dynamic contexts.

Recent examples of impressive cooperative behaviors achieved by MARL, such as outperforming humans in real-time strategy games (Berner et al. 2019; Vinyals et al. 2019) and air combat simulations (Pope et al. 2021), represent remarkable advancements. Developing agents for dynamic environments, however, often requires prohibitive computational resources, limiting practical applications (Du and Ding 2021). To address these challenges, researchers have explored integrating domain knowledge to enhance training effectiveness. This can be achieved by incorporating human replay data (Berner et al. 2019) or employing hierarchical learning based on prior knowledge (Pope et al. 2021). These approaches aim to improve the training of the agents' policy models, which are crucial for effective decision-making in dynamic contexts. Over time, these models have evolved from tabular-based solutions to neural networks and attention-based approaches. Attention-based policies, following their success in natural language processing (Vaswani et al. 2017), are now expanding the possibilities in MARL by enabling more efficient and effective integration of domain knowledge into the learning process.

The main contribution of this paper is a novel attention-based method that integrates domain knowledge at the action level of the MARL process. This eliminates the need for agents to learn low-level skills from scratch. In standard MARL scenarios, our method achieves state-of-the-art results while requiring only 20% of the training effort compared to traditional approaches. Furthermore, our approach demonstrates significant scalability and adaptability advantages, maintaining its performance with varying numbers of agents and observation sizes.

## Background and Related Work

### Domain Knowledge in the MARL Process

Integrating domain knowledge into MARL offers promising solutions, as demonstrated by Zheng incorporating industrial knowledge graphs (Zheng et al. 2021) and Yu optimizing task scheduling in human-robot systems (Yu, Huang, and Chang 2021). The complexity of specialist knowledge, however, often conflicts with the data-driven nature of MARL systems (Vinyals et al. 2019). Hierarchical Learning, which subdivides problems to simplify the learning process, is an interesting alternative to integrating experts' knowledge into model structures. This type of strategy yields impressive results in air combat (Pope et al. 2021), autonomous driving (Wang et al. 2020), and real-time strategic games (Berner et al. 2019). Han and Nguyen also directly integrate rule-based human behavior to optimize the exploration capability of the MARL process (Han et al. 2020; Nguyen, Nguyen, and Nahavandi 2019).

Our methodology draws inspiration from hierarchical learning by encapsulating domain knowledge into dynamically generated higher-level tasks. Like Han and Nguyen, our model utilizes this knowledge to guide exploration. Our approach, however, differs as tasks become an integral part of the model behavior, introducing a deliberate bias to steer agent behavior while adapting to possible limitations created by the embedded knowledge. This intentional bias is designed to steer agent behavior in specific directions, leveraging the flexibility and power of the attention-based policy to find new pathways for effective collaborative behavior.

### Attention-based Policy

Attention-based Policy represents an alternative approach that primarily utilizes the attention mechanism (Niu, Zhong, and Yu 2021) as the core component of its learning models. Shifts from traditional neural networks to adaptive models like Graph Neural Networks (GNNs) and Attention-based models are gaining relevance in MARL. The incorporation of attention mechanisms, inspired by models such as Transformers (Vaswani et al. 2017), is emerging as a potent tool for enhancing model robustness and scalability, as demonstrated by various studies (Iqbal and Sha 2019; Pu et al. 2022; Zhang, Yang, and Başar 2021; Omee, Hossain, and Hossain 2021; Sethi et al. 2021; Xiao, Yuan, and Wang 2023). The ability of attention-based models to process variable-length inputs and offer strong insights into extensive contexts is highly relevant for MARL, particularly when compared to the more commonly used MLP-based policy models. In our work, the dynamic nature of task generation positions attention models as forward-thinking solutions, enhancing the capacity to learn complex, context-based decisions derived from a varying number of tasks.

## Proposed Methodology

The proposed methodology enhances MARL efficiency by embedding prior knowledge into higher-level tasks, simplifying the learning process and focusing on collaboration. The approach aims to be easily integrable with well-established MARL solutions without significant structural changes. Figure 1 provides a general view of the integration of the Task Generator, the Attention-based Policy and the Task to Action Converter to a traditional MARL process.

### Task Generator

The Task Generator is designed to process inputs from environmental observations and construct potential tasks at each step. The methodology for creating these tasks is deeply connected to the problem being addressed and relies on the integration of human expertise. This involves identifying problem knowledge that can be encapsulated into higher-level tasks, aimed at simplifying the learning process while retaining enough flexibility to foster the emergence of collaborative behaviors. Each task is formulated based on data received from the environment and is defined by a set of numerical features. Each type of task, classified by its embedded knowledge, is assigned a unique code using one-shot representation. These codes enable the policy to identify the types of tasks during the learning process. All tasks are sent to the policy as a sequence of vectors, each vector carrying the features of one possible task. It is expected that collectively, these tasks carry sufficient information to provide the policy with a thorough understanding of the current state.

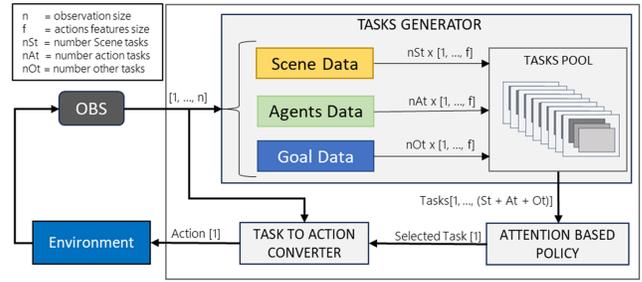

Figure 1: Attention Task-based action selection for MARL.

### The Attention-Based Policy

The Attention-Based Policy, the core of the learning process, interprets tasks from the Task Generator to select the best option. Built on a Multi-Head Attention (MHA) architecture (Vaswani et al. 2017), it consists of four main components. An encoder first processes each task, expanding its dimensionality via an MLP with three 64-node hidden layers. The enhanced tasks then feed into two connected MHAs, followed by another MLP with two 32-node hidden layers to generate a single output value for each task. The result is a vector indicating the relevance of each task. Skip connections and normalization layers are integrated after each MHA. The selected task is then processed by the Task to Action Converter generating the agent's final action.

### Task to Action Converter

The Task to Action Converter encapsulates the expertise needed to determine the appropriate lower-level action corresponding to the selected task. This process is intricately linked to the rationale behind the task's creation and is designed to output the best-known action based on its embedded knowledge for that situation. The converter can utilize either the features of the selected task or all the information available through the observation to decide the next action.

## The TestBed

To validate the proposed methodology, we select the MPE and SISL, two established scenarios focusing on learning collaborative behavior. PettingZoo (Terry et al. 2021), version 1.24.0, serves as the environment testbed. The MPE Simple Spread is a scenario where agents must efficiently occupy landmarks while avoiding collisions, testing cooperative and spatial-awareness skills. The SISL Pursuit is a predator-prey game in a discrete setting, where agents work together to trap the prey with limited views and random prey moves. We develop extended versions of the base environments, including the Task Generator and the Task to Action Converter, to process observations, generate tasks, and produce valid actions the policy selection. Table 1 detail the tasks for the Single Spread and Pursuit, including embedded knowledge, action conversion, and selected features.

| Type | Embedded Knowledge | Action Conversion | Features |
|---|---|---|---|
| MPE - Go Landmark | The primary objective is to get closer to a landmark. | Use a PID controller to convert distance and angle from the landmark into discrete actions. | [1, 0, distance, sin(radial), cos(radial), own_velocity] |
| MPE - Avoid Collision | Collisions result in penalties and propel you to an unexpected position. | Select the closest action to the perpendicular direction of the ally's direction. | [0, 1, ally_distance, sin(ally_radial), cos(ally_radial), own_velocity] |
| SISL - Explore | In specific situations, it is better to move in a selected direction to find better conditions. | Direct conversion, as base actions are already directions. | [1, 0, 0, unitary_distance, sin(direction), cos(direction), own_agent_velocity] |
| SISL - Pursuer Evader | The objective is to capture evaders, necessitating movement towards them. Additionally, it involves splitting up for effective capture. | Select action that minimizes distance to the evader and facilitates splitting from allies when close. | [0, 1, 0, evader_distance, sin(evader_radial), cos(evader_radial), own_agent_velocity] |
| SISL - Follow Ally | Working alone is not feasible, and allies may have a better understanding of the situation. | Select action that minimizes distance to the allies. | [0, 0, 1, ally_distance, sin(ally_radial), cos(ally_radial), own_velocity] |

Table 1: MPE Simple Spread and SISL Pursuit Task-Based Actions definition and explanation

## Experiments

Despite efforts towards standardization with well-defined environments like MPE and SISL the current landscape of MARL studies often lacks clarity in defining state-of-the-art results and baselines (Gorsane et al. 2022). For example, in MPE's Simple Spread benchmarks, (Papoudakis et al. 2021) identifies QMIX as the best algorithm with a mean reward of -126.3, while (Hu et al. 2023) reports significantly lower performance for QMIX, with a best mean reward of -189.2. In SISL environments, (Gerstgrasser et al. 2023) favor an Independent Learning DQN-based algorithm named SUPER, whereas (Siu, Traish, and Da Xu 2021) propose a value decomposition method called GUM, but direct comparisons are challenging due to different infrastructures and evaluation conditions. Initiatives like MARLLib (Hu et al. 2023), BenchMARL (Bettini, Prorok, and Moens 2023), and Tianshou (Zheng et al. 2021) are pushing for frameworks that enable standardized evaluation and fair comparison in MARL. In this work, we adopt these frameworks to provide a more consistent and comparative analysis of MARL algorithms.

To showcase the benefits of our approach, we conduct experiments to define desired performances, comparable to published results. We also establish a random action-based baseline to ensure minimal learning effectiveness and serve as a reference for comparing different environments. Baselines are calculated from 30 repetitions of 50 episodes, choosing random uniformly distributed actions at each step.

For the Simple Spread environment (single_spread_v3), we use BenchMARL to run seven MARL algorithms: IPPO, IQL, ISAC, MAPPO, MASAC, QMIX, and VDN (Bettini, Prorok, and Moens 2023), analyzing their performance as initial reference. For the Pursuit environment (pursuit_v4), we evaluate a variation of the algorithm from (Gerstgrasser et al. 2023) and compare maximum rewards with previous studies. We analyze learning evolution and compare the best mean reward of 50 episodes from test phases, applying the Bootstrap 95% Confidence Interval (CI) (Gorsane et al. 2022) to validate statistical differences.

To demonstrate the scalability and adaptability advantages of our approach, we evaluate it under two different conditions for the Single Spread and Pursuit environments, where traditional solutions are expected to struggle. In the Single Spread scenario, a policy initially trained with three agents was tested in new scenarios involving 6, 9, 12 and 15 agents and landmarks, without any additional training. Since the observation includes all landmarks and allies' information, changing the size of the scenario will change the observation size. For the Pursuit environment, we assess the model's adaptability to perform in varying field of view (FOV) sizes after training for only one of them.

## Results and Discussion

### Random Action Baselines

Our initial Simple Spread evaluations indicate notable inconsistencies across different environments. For the random action approach, BenchMARL and Tianshou yield a mean reward close to -27, compared to -117 for MARLLib. An investigation reveals that MARLLib uses an outdated version of the MPE Simple Spread, which contains a critical error. This error leads to all agents erroneously receiving a -1.0 penalty at each step, falsely simulating self-collisions. Intriguingly, this issue appears overlooked in many published studies. Additionally, the reward in MARLLib is the sum of all agents' rewards; in BenchMARL and Tianshou, it is the mean. Correcting these differences, the baseline results become consistent among the three solutions with a mean reward of -27.1 (CI: -27.8, -26.3). Regarding the SISL Pursuit, we run evaluations using Tianshou with the baseline mean reward as 368.1 (CI: 371.7, 364.3) for random actions.

### Benchmark Results

The evaluation of the Single Spread scenario to establish a benchmark yields varied results. Among the algorithms, four (IPPO, IQL, VDN, and ISAC) demonstrate comparable performances, while three (MAPPO, MASAC and QMIX) fail to find effective policies. This underperformance can be attributed to the use of default parameters without fine-tuning. The results of the top performing four algorithms, however, align well with leading results reported in the literature, validating their use as a reliable benchmark. The highest mean reward is recorded by IPPO at -8.67 (CI: -9.48, -7.90) in iteration 425, which is even the best-known result for the adopted configuration. We also calculate the mean reward for the tests in the last 50 steps, with the best result for IPPO being -9.81 (CI: -9.94, -9.67). Regarding the SISL Pursuit, the benchmark achieves a best reward of 633.6 (CI: 597.2, 668.3) which aligns with the best results in the literature. We also watch simulations of the best policies and confirm visually that the agents collaborate in a very efficient manner.

### Task-Based Action Results

In the Simple Spread, our methodology outperforms the benchmarks, achieving a notable best mean reward of -7.23 (CI: -7.74 to -6.73), which surpasses the benchmark by 16.6%. The mean for the last 50 episodes is -7.75 (CI: -7.85 to -7.66) which is 20.1% better than the benchmark. Similarly, the Pursuit scenario showcases the efficiency of our task-based action approach. The model achieves the highest mean reward of 673.7 (CI: 643.3, 703.1), statistically equivalent to the benchmark when considering the overlap within the 95% CIs. Figure 2 plots the training evolution, demonstrating the greater efficiency of our approach by achieving the previous best result in fewer than 80 training steps for the Single Spread and less than 40 in the Pursuit.

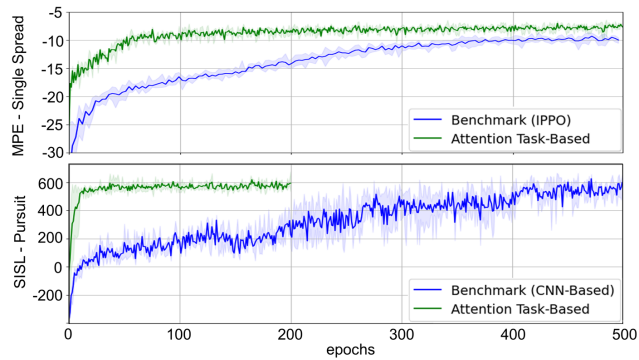

Figure 2: Attention-driven task-based solution Mean Rewards for MPE Single Spread and SISL Pursuit.

### Scalability Results and Adaptability

In evaluating scalability within the MPE Single Spread scenario, our approach has proven its robustness by effectively handling larger scenarios without any new training, achieving superior performance compared to the benchmark, which requires new training for each case. Figure 3 illustrates these results and how the number of agents impacts the adjusted mean reward. The adjusted mean reward, calculated as the default episode reward divided by the number of agents, indicates the complexity added by an increased number of agents, not merely due to a higher number of components. The results from the random actions highlight a significant increase in penalties as the number of agents grows, evidenced by a linear decline in adjusted rewards.

Our model's adaptability in the SISL Pursuit scenario is highlighted by its reliable performance across variable FOV sizes. After training only with a 7x7 FOV, our approach achieves a mean reward of 587.3 (CI: 557.5, 615.5) with a 5x5 FOV, while the benchmark achieves 491.0 (CI: 439.3, 541.5) even after new training. For a 9x9 FOV, our results are statistically equivalent to the benchmark, achieving 605.3 (CI: 569.30, 639.99) with the benchmark at 569.8 (CI: 521.3, 617.5). Our results, however, are obtained without any new training. Interestingly, the increased FOV did not lead to an improvement in the benchmark's performance, possibly due to the increased complexity introduced by the larger FOV.

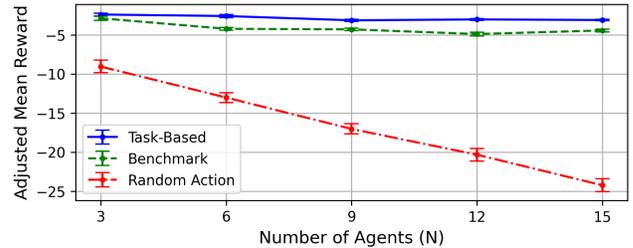

Figure 3: MPE Single Spread scalability with 95% CI.

## Conclusion and Future Work

This study introduces an alternative MARL approach, integrating domain knowledge through higher-level tasks and attention-based policy. Experiments across standard MARL scenarios demonstrate our methodology's potential, surpassing benchmarks in learning efficiency and effectiveness. The inherent flexibility enables scalable adaptation to more complex scenarios and variable observations while preserving cooperative capabilities without additional training. Our method relies on designing higher-level tasks, which can either limit or bolster the model's potential. Personalizing tasks to specific problems underscores the need for customization in each application, which is not always feasible. When customization is feasible, however, it can significantly simplify the overall learning process. Future work will explore our methodology's benefits in more realistic settings, such as autonomous vehicles and military operations, to further assess its potential in real-world applications.

## Acknowledgments

This work was supported by the National Council for Scientific and Technological Development—CNPq 307691/2020-9 and the Brazilian Air Force Postgraduate Program in Operational Applications (PPGAO).

# References

Berner, C.; Brockman, G.; Chan, B.; Cheung, V.; Debiak, P.; Dennison, C.; Farhi, D.; Fischer, Q.; Hashme, S.; Hesse, C.; et al. 2019. Dota 2 with large scale deep reinforcement learning. *arXiv preprint arXiv:1912.06680*.

Bettini, M.; Prorok, A.; and Moens, M.-F. 2023. Benchmarl: Benchmarking multi-agent reinforcement learning. *arXiv preprint arXiv:2006.07869*.

Du, W., and Ding, S. 2021. A survey on multi-agent deep reinforcement learning: from the perspective of challenges and applications. *Artificial Intelligence Review* 54(5):3215–3238.

Gerstgrasser, M.; Danino, T.; Keren, S.; and Paulson, J. A. 2023. Selectively sharing experiences improves multi-agent reinforcement learning. *arXiv preprint arXiv:2311.00865*.

Gorsane, R.; Mahjoub, O.; de Kock, R. J.; Dubb, R.; Singh, S.; and Pretorius, A. 2022. Towards a standardised performance evaluation protocol for cooperative marl. In *Advances in Neural Information Processing Systems*, volume 35, 5510–5521.

Han, X.; Tang, H.; Li, Y.; Kou, G.; and Liu, L. 2020. Improving multi-agent reinforcement learning with imperfect human knowledge. In *Lecture Notes in Computer Science (including subseries Lecture Notes in Artificial Intelligence and Lecture Notes in Bioinformatics)*, volume 12397, 369–380. Springer.

Hu, S.; Zhong, Y.; Gao, M.; Wang, W.; Dong, H.; Liang, X.; Li, Z.; Chang, X.; Yang, Y.; and Fuxi Lab, N. A. 2023. Marllib: A scalable and efficient library for multi-agent reinforcement learning. *Journal of Machine Learning Research* 24:1–23.

Iqbal, S., and Sha, F. 2019. Actor-attention-critic for multi-agent reinforcement learning. In *International conference on machine learning*, 2961–2970. PMLR.

Nguyen, T.; Nguyen, N. D.; and Nahavandi, S. 2019. Multi-agent deep reinforcement learning with human strategies. In *Proceedings of the IEEE International Conference on Industrial Technology*, 1357–1362. IEEE.

Niu, Z.; Zhong, G.; and Yu, H. 2021. A review on the attention mechanism of deep learning. *Neurocomputing* 452:48–62.

Omee, F.; Hossain, M. Z. I.; and Hossain, E. 2021. Multi-agent reinforcement learning for resource allocation in iot networks. *IEEE Access* 9:93533–93546.

Papoudakis, G.; Christianos, F.; Schäfer, L.; and Albrecht, S. V. 2021. Benchmarking multi-agent deep reinforcement learning algorithms in cooperative tasks. *Thirty-fifth Conference on Neural Information Processing Systems Datasets and Benchmarks Track (Round 1)*.

Pope, A. P.; Ide, J. S.; Micovic, D.; Diaz, H.; Rosenbluth, D.; Ritholtz, L.; Twedt, J. C.; Walker, T. T.; Alcedo, K.; and Javorsek, D. 2021. Hierarchical reinforcement learning for air-to-air combat. In *2021 International Conference on Unmanned Aircraft Systems, ICUAS 2021*, 275–284. IEEE.

Pu, Z.; Wang, H.; Liu, Z.; Yi, J.; and Wu, S. 2022. Attention enhanced reinforcement learning for multi agent cooperation. *IEEE Transactions on Neural Networks and Learning Systems*.

Sethi, K.; Madhav, Y.; Kumar, R.; and Bera, P. 2021. Attention based multi-agent intrusion detection systems using reinforcement learning. *Journal of Information Security and Applications* 61:102923.

Siu, C.; Traish, J.; and Da Xu, R. Y. 2021. Greedy unmixing for q-learning in multi-agent reinforcement learning. *arXiv preprint arXiv:2109.09034*.

Terry, J.; Black, B.; Grammel, N.; Jayakumar, M.; Hari, A.; Sullivan, R.; Santos, L.; Dieffendahl, C.; Horsch, C.; Perez-Vicente, R.; et al. 2021. Pettingzoo: Gym for multi-agent reinforcement learning. In *Advances in Neural Information Processing Systems*, volume 34, 15032–15043.

Vaswani, A.; Brain, G.; Shazeer, N.; Parmar, N.; Uszkoreit, J.; Jones, L.; Gomez, A. N.; Kaiser, L.; and Polosukhin, I. 2017. Attention is all you need. In *Advances in Neural Information Processing Systems*, volume 30.

Vinyals, O.; Babuschkin, I.; Czarnecki, W. M.; Mathieu, M.; Dudzik, A.; Chung, J.; Choi, D. H.; Powell, R.; Ewalds, T.; Georgiev, P.; et al. 2019. Grandmaster level in starcraft ii using multi-agent reinforcement learning. *Nature* 575(7782):350–354.

Wang, J.; Wang, Y.; Zhang, D.; Yang, Y.; and Xiong, R. 2020. Learning hierarchical behavior and motion planning for autonomous driving. In *IEEE International Conference on Intelligent Robots and Systems*, 2235–2242. IEEE.

Xiao, J.; Yuan, G.; and Wang, Z. 2023. A multi-agent flocking collaborative control method for stochastic dynamic environment via graph attention autoencoder based reinforcement learning. *Neurocomputing* 126379.

Yu, T.; Huang, J.; and Chang, Q. 2021. Optimizing task scheduling in human-robot collaboration with deep multi-agent reinforcement learning. *Journal of Manufacturing Systems* 60:287–297.

Zhang, K.; Yang, Z.; and Başar, T. 2021. Multi-agent reinforcement learning: A selective overview of theories and algorithms. In *Handbook of reinforcement learning and control*. Springer. 321–384.

Zheng, P.; Xia, L.; Li, C.; Li, X.; and Liu, B. 2021. Towards self-x cognitive manufacturing network: An industrial knowledge graph-based multi-agent reinforcement learning approach. *Journal of Manufacturing Systems* 60:373–382.